\documentclass[journal]{IEEEtran}

\usepackage{cite}
\usepackage[pdftex]{graphicx}
\usepackage{dblfloatfix}
\usepackage{setspace}
\usepackage{siunitx}
\usepackage{algorithm}
\usepackage{algpseudocode}
\algrenewcommand\algorithmicrequire{\textbf{Input}:}
\algrenewcommand\algorithmicensure{\textbf{Output}:}
\algnewcommand{\LineComment}[2]{\Statex \hspace{#1} \(\triangleright\) #2}
\algnewcommand{\IfThen}[2]{
  \State \algorithmicif\ #1\ \algorithmicthen\ #2}

\usepackage{array}
\usepackage{multirow}
\usepackage{booktabs}
\usepackage{tabularx}

\newcolumntype{C}[1]{>{\centering\arraybackslash}p{#1}}
\newcolumntype{L}[1]{p{#1}}
\newcolumntype{Y}{>{\centering\arraybackslash}X}

\usepackage{amsmath}
\usepackage{amsfonts}
\usepackage{amsthm}
\usepackage{amssymb}
\usepackage{multidef}
\usepackage{mathtools}
\multidef[prefix=set]{\ensuremath{\mathbb{#1}}}{A-Z}

\multidef[prefix=cal]{\ensuremath{\mathcal{#1}}}{A-Z}

\multidef[prefix=v]{\ensuremath{\mathbf{#1}}}{a-z}
\newcommand{\vell}{\boldsymbol{\ell}}

\multidef[prefix=m]{\ensuremath{\mathbf{#1}}}{A-Z}
\newcommand{\mLambda}{\boldsymbol{\Lambda}}
\newcommand{\mDelta}{\boldsymbol{\Delta}}
\newcommand{\mSigma}{\boldsymbol{\Sigma}}
\newcommand{\ones}{\boldsymbol{1}}
\newcommand{\zeros}{\boldsymbol{0}}

\newcommand{\upper}{\mathrm{upper}}
\newcommand{\diag}{\mathrm{diag}}
\newcommand{\trace}{\mathrm{tr}}
\DeclareMathOperator*{\minimize}{minimize}
\DeclarePairedDelimiter\norm{\lVert}{\rVert}

\DeclareMathOperator*{\argmin}{argmin}

\theoremstyle{plain}

\newtheorem{lemma}{Lemma}
\newtheorem{theorem}{Theorem}

\theoremstyle{definition}
\newtheorem{definition}{Definition}

\theoremstyle{remark}
\newtheorem{remark}{Remark}

\begin{document}

\title{Signed Graph Learning: Algorithms and Theory}
\author{Abdullah Karaaslanli$^{\dagger}$,\ %
Bisakh Banerjee$^{\dagger}$,\ %
Tapabrata Maiti and Selin Aviyente%
\thanks{This work was supported by the National Science Foundation in part by CCF-2211645 and CCF-2312546.}%
\thanks{$\dagger$ The first two authors contributed equally to this manuscript.}%
\thanks{A.K. and S.A are with the Department of Electrical and Computer Engineering, Michigan State University, East Lansing, MI 48824 USA. B.B. and T.M. are with the Department of Statistics and Probability, Michigan State University, East Lansing, MI 48824 USA. (e-mail: karaasl1@msu.edu; banerj40@msu.edu; maiti@msu.edu; aviyente@egr.msu.edu).}%
\thanks{This work has been submitted to the IEEE for possible publication.
Copyright may be transferred without notice, after which this version may
no longer be accessible.}%
}

\maketitle

\begin{abstract}
Real-world data is often represented through the relationships between data samples, forming a graph structure. In many applications, it is necessary to learn this graph structure from the observed data. Current graph learning research has primarily focused on unsigned graphs, which consist only of positive edges. However, many biological and social systems are better described by signed graphs that account for both positive and negative interactions, capturing similarity and dissimilarity between samples. In this paper, we develop a method for learning signed graphs from a set of smooth signed graph signals. Specifically, we employ the net Laplacian as a graph shift operator (GSO) to define smooth signed graph signals as the outputs of a low-pass signed graph filter defined by the net Laplacian. The signed graph is then learned by formulating a non-convex optimization problem where the total variation of the observed signals is minimized with respect to the net Laplacian. The proposed problem is solved using alternating direction method of multipliers (ADMM) and a fast algorithm reducing the per-ADMM iteration complexity from quadratic to linear in the number of nodes is introduced. Furthermore, theoretical proofs of convergence for the algorithm and a bound on the estimation error of the learned net Laplacian as a function of sample size, number of nodes, and graph topology are provided. Finally, the proposed method is evaluated on simulated data and gene regulatory network inference problem and compared to existing signed graph learning methods. 
\end{abstract}

\section{Introduction}
\label{sec:introduction}
Many machine learning and signal processing problems include datasets where samples are dependent to each other. These relations are often modeled and studied with graphs, where samples and interactions are represented as nodes and edges, respectively \cite{dong2020graph, newman2018networks}. While structure of the graph is known in some applications, such as interactions between people in a social network or roads connecting intersections in a traffic network are observable; numerous applications lack readily available graphs. For example, the functional connectivity between brain regions \cite{sporns2012discovering} and regulatory interactions between genes and proteins \cite{karlebach2008modelling} are not directly observable. Effective data processing in these crucial applications requires inference of the graph topology \cite{dong2019learning, mateos2019connecting}.

Various methods have been developed for \textit{graph learning} problem where the topology of the unknown graph is learned from observed nodal data, also known as \textit{graph signals}. The problem is considered in different domains, including statistics \cite{drton2017structure}, graph signal processing (GSP) \cite{dong2019learning, mateos2019connecting}, and graph neural networks (GNN) \cite{kipf2018neural}. Statistical methods, such as graphical lasso \cite{friedman2008sparse}, are usually based on Gaussian graphical models, where the aim is to learn the precision matrix representing conditional dependencies between nodes. Methodologies using GSP define the relation between graph signals and graph topology using signal processing concepts, such as stationarity \cite{thanou2017learning, segarra2017network, pasdeloup2017characterization}, non-stationarity \cite{shafipour2021identifying}, and smoothness \cite{dong2016learning, kalofolias2016learn, berger2020efficient}. Finally, GNN-based techniques use graph convolutional networks and the more general message passing networks for relational inference \cite{kipf2018neural, franceschi2019learning, chen2020iterative}.

Most existing graph learning methods have been limited to unsigned graphs, i.e., graphs with positive edges encoding similarity between node pairs. However, in real-world scenarios, there exists pairwise dissimilarity among data samples, which are more appropriately modeled using negative edges. The resulting graph in such cases is signed, where nodes are connected with either positive or negative edges indicating similarity and dissimilarity of the node pairs, respectively \cite{dittrich2020signal}. For instance, international relationships among countries can be represented as a signed graph, with each vertex representing a country \cite{chu2016finding}. Friendly countries are connected by positive edges, while hostile countries are connected by negative edges. Similarly, in online social networks \cite{girdhar2017signed}, interactions between individuals can be classified into two categories: trust versus distrust and friendly versus antagonistic. In biological networks such as gene regulatory networks (GRNs), interactions between genes can be activating or inhibitory, leading to both positive and negative edges \cite{badia2023gene}.

Recently, smoothness-based unsigned graph learning techniques from GSP literature have been adapted for signed graph learning problem. For unsigned graphs, these approaches employ the Laplacian matrix as the graph shift operator (GSO) in order to define smooth unsigned graph signals as the outputs of a low-pass graph filter \cite{dong2016learning, kalofolias2016learn}. An unsigned graph is then learned by minimizing the quadratic form of the Laplacian, which quantifies the total variation of signals, with respect to the Laplacian matrix. Extension of smoothness to signed graph learning is introduced in \cite{matz2020learning}, where the signed Laplacian \cite{kunegis2010spectral} is utilized as the GSO. The quadratic form of signed Laplacian is minimized with respect to the signed Laplacian to learn the unknown signed graph. Small values of the signed Laplacian quadratic form implies that signal values across positive edges are similar and signal values across the negative edges are dissimilar. However, dissimilarity in this framework is modeled as signal values having opposite signs, which is not suitable when graph signals are either all positive- or negative-valued. For example, gene expression data is all non-negative as it measures gene counts in cells or tissues \cite{chen2019single}. Learning the signed Laplacian in such cases results in negative edges not being inferred properly.

In order to address this shortcoming with signed Laplacian, \cite{karaaslanli2022scsgl} formulates signed graph learning problem by assuming that observed graph signals have small total variation over positive edges and large total variation over negative edges. In this manner, the problem of learning a signed graph is posed as learning two unsigned graphs constructed from positive and negative parts of the signed graph. The Laplacian matrices of these two unsigned graphs are then learned by minimizing the variation over the positive edges while maximizing the variation over the negative edges within a single optimization problem. The process of splitting the signed graph into two unsigned graphs introduces complementarity constraint resulting in a non-convex optimization framework. To eliminate this constraint, \cite{fong2024efficient} reconsiders the formulation in \cite{karaaslanli2022scsgl} by learning the adjacency matrix of the signed graph. The resulting optimization problem is convex and solved with proximal alternating direction method of multipliers (pADMM) with global convergence and a local linear convergence rate. In \cite{yokota2025efficient}, a linear programming based framework is introduced for learning the balanced signed graph Laplacian directly from data. However, this framework does not provide smoothness of the observed data with respect to the learned Laplacian and its goal is to impose balance to learn the signed graph, which is beyond the scope of this paper. 

In this paper, we revisit our previous approach in \cite{karaaslanli2022scsgl} by reconsidering its formulation from a GSP perspective, while providing a new optimization procedure and theoretical analysis. In particular, we first define smoothness for signed graphs by employing an appropriate GSO for signed graphs, i.e., \textit{the net Laplacian} \cite{stanic2020spectrum}, which also addresses the shortcomings with the previous GSO operator (the signed Laplacian) used for signed graphs. A smooth signed graph signal is introduced as the output of a low-pass graph filter defined through the net Laplacian. The optimization problem in \cite{karaaslanli2022scsgl} can then be derived by minimizing the quadratic form of the net Laplacian, which provides a measure of smoothness for the observed signed graph signals. We provide an alternating direction method of multipliers (ADMM) based solution to this optimization problem and provide theoretical guarantees for its convergence to a local minimum. We also propose a fast implementation to reduce per-ADMM time complexity from quadratic to linear in the number of nodes. Furthermore, we present theoretical results regarding the estimation error of the learned net Laplacian in relation to the number of observed signals and nodes. The proposed method is evaluated against existing signed graph learning methods using both simulated and real data.

\section{Background}
\label{sec:background}

\subsection{Notations}
\label{ssec:background-notations}

Lowercase and uppercase letters ($n$ or $N$), lowercase bold letters ($\vx$) and uppercase bold letters ($\mX$) are used to represent scalars, vectors and matrices, respectively. 
For a vector $\vx$, $x_{i}$ is its $i$th element; while for a matrix $\mX$, $X_{ij}$ is its $ij$th element. $i$th row and column of a matrix $\mX$ are shown as $\mX_{i\cdot}$ and $\mX_{\cdot i}$, respectively and they are both assumed to be column-vectors. All-one vector, all-zero vector and identity matrix are are shown as $\mathbf{1}$, $\mathbf{0}$, and $\mI$. $\trace(\cdot)$, $^\dagger$ and $\odot$ are utilized to refer to the trace operator, pseudo-inverse of a matrix, and Hadamard product. $\diag(\cdot)$ operator is defined such that for a vector, $\diag(\vx) = \mX$ where $\mX$ is a diagonal matrix with $X_{ii} = x_i$; for a matrix, $\diag(\mX) = \vx$ where $x_i = X_{ii}$. For a set $\calS$, $|\calS|$ depicts its cardinality, i.e., the number of elements in $\calS$. 

For a matrix $\mX \in \mathbb{R}^{n \times m}$, $\norm{\mX}_{2}$, $\norm{\mX}_{F}$ and $\norm{\mX}_{1}$ represent its spectral, Frobenius and element-wise $\ell_1$ norm, respectively. 
For vectors $\boldsymbol{y}_n, \boldsymbol{x}_n \in \mathbb{R}^p$, the notation $\boldsymbol{y}_n \asymp \boldsymbol{x}_n$ means that $\boldsymbol{y}_n = \mathcal{O}(\boldsymbol{x}_n)$ and $\boldsymbol{x}_n = \mathcal{O}(\boldsymbol{y}_n)$, implying the existence of a constant $0 < M < \infty$ such that $\|\boldsymbol{x}_n\| \leq M \|\boldsymbol{y}_n\|$ for all $n \geq 1$. $\boldsymbol{y}_n = o(\boldsymbol{x}_n) \quad \text{as } n \to \infty$, if  $ \frac{ \| \boldsymbol{y}_n \| }{ \| \boldsymbol{x}_n \| } \rightarrow 0$.    For random vectors $\boldsymbol{y}_n, \boldsymbol{x}_n \in \mathbb{R}^p$, the notation $\boldsymbol{y}_n = \mathcal{O}_P(\boldsymbol{x}_n)$ means that for any $\varepsilon > 0$, there exists a constant $0 < M < \infty$ such that $\mathbb{P}(\|\boldsymbol{y}_n\| \leq M \|\boldsymbol{x}_n\|) \geq 1 - \varepsilon$ for all $n \geq 1$. %

\subsection{Graphs}
\label{ssec:background-graphs}

A graph can be defined by the tuple $G=(V, E)$ where $V$ is the node set with $|V| = n$ and $E$ is the edge set with elements $e_{ij}$ representing the interaction between nodes $i$ and $j$. $G$ is an undirected graph, if $e_{ij} \in E$ implies $e_{ji} \in E$. In this paper, all graphs are assumed to be undirected. Edges of a graph are associated with edge weights, $w_{ij} > 0$, and edge signs, $s_{ij} \in \{-1, 1\}$. $G$ is \textit{unsigned} when $s_{ij} = 1, \forall e_{ij} \in E$ and \textit{signed}, otherwise. Adjacency matrix of an undirected graph $G$ is a symmetric matrix $\mA \in \setR^{n \times n}$ with elements $\mA_{ij} = s_{ij}w_{ij}$ if $e_{ij} \in E$ and $\mA_{ij} = 0$, otherwise. For signed graphs, $\mA$ can be decomposed based on edge signs as $\mA = \mA^+ - \mA^-$ where $A_{ij}^+ = {\rm max}(s_{ij}, 0)w_{ij}$ and $A_{ij}^- = |{\rm min}(s_{ij}, 0)|w_{ij}$. 

\subsection{Graph Signal Processing}
\label{ssec:background-gsp}

A graph signal defined on an \textit{unsigned} graph $G$ is a vector $\vx \in \setR^n$ where $x_i$ is the signal value of node $i$. The goal of GSP is to analyze $\vx$ by incorporating the structure of $G$ into the analysis. This is done by representing $G$ by a graph shift operator, which is a symmetric matrix $\mS \in \setR^{n \times n}$ where $S_{ij} \neq 0$ if and only if $e_{ij} \in E$ or $i = j$. A commonly used GSO is the Laplacian matrix $\mL = \diag(\mA \ones) - \mA$, whose eigendecomposition is $\mL = \mV \mLambda \mV^\top$ where $\mV = [\vv_1, \dots, \vv_n]$ contains eigenvectors and $\mLambda$ is the diagonal matrix of eigenvalues $0 = \lambda_1 \leq \lambda_2 \leq \dots \leq \lambda_n$. Eigendecomposition of $\mL$ can be used to derive graph Fourier transform (GFT) by associating its eigenvalues to a notion of frequency similar to classical signal processing \cite{shuman2013emerging}. For a graph signal $\vx$, its Dirichlet energy with respect to $G$ is calculated using the quadratic form of $\mL$:
\begin{align}
    \label{eq:smoothness-node-domain}
    \vx^\top \mL \vx = \frac{1}{2}\sum_{i \neq j}^n A_{ij} (x_{i} - x_{j})^2,
\end{align}

\noindent which quantifies the total variation of $\vx$ such that smaller values indicate $\vx$ has low oscillation across edges, i.e., signal values on strongly connected nodes are similar. Since $\vv_i^\top \mL \vv_i = \lambda_i$ for any eigenvector $\vv_i$, eigenvalues and eigenvectors of $\mL$ provide a notion of frequency where smaller eigenvalues correspond to low-frequencies \cite{shuman2013emerging}. GFT of $\vx$ is then defined as $\widehat{\vx} = \mV^\top \vx$ with the inverse GFT defined as: $\vx = \mV\widehat{\vx} = \sum_{i=1}^n \widehat{x}_i \vv_i$ where $\widehat{x}_i$ is the $i$th graph Fourier coefficient which quantifies how much of $\vx$'s energy lies at graph frequency component $\vv_i$. A fundamental tool for modeling graph signals with respect to the underlying graph structure is graph filters \cite{isufi2024graph}. A graph filter is a matrix $\mH \in \setR^{n \times n}$ constructed as a function of GSO, i.e., $\mH = h(\mL) = \mV h(\mLambda) \mV^\top$. Given an input signal $\vx_0$, graph filtering can then be used to model $\vx$ as $\vx = \mH \vx_0$. In the graph Fourier domain:
\begin{align}
    \label{eq:filtering-freq}
    \vx = \mV h(\mLambda) \mV^\top \vx_0 = \sum_{i=1}^n h(\lambda_i) \widehat{x}_{0i} \vv_i,
\end{align}

\noindent where $h(\lambda_i)$ is the frequency response of the graph filter. Akin to classic signal processing, $h(\lambda_i)$ can be low-pass, high-pass or band-pass, which attenuates or amplifies specific Fourier coefficients of the input signal. 

\subsection{Learning the Graph Structure from Smooth Signals}
\label{eq:background-unsigned-gl}

Given a set of $m$ graph signals $\calX = \{\vx_i\}_{i=1}^m$ that are defined on an \textit{unknown} unsigned graph $G$, graph learning infer the structure of $G$ from $\calX$ by modeling the relationship between $G$ and $\calX$. This paper considers the case where $\vx_i$'s are the outputs of a low-pass graph filter, which preserves the low-frequency content of its excitation signal while filtering out the high-frequency content. A low-pass graph filter generates smooth graph signals, i.e., they have low total variation with respect to the underlying graph structure. Given a set of smooth graph signals, $G$ is then learned by minimizing the total variation of signals with respect to $\mL$ \cite{dong2016learning}:
\begin{align}
\label{eq:unsigned-gl}
\begin{aligned}
    \minimize_{\mL} &\ \trace({\mX}^\top \mL \mX) + \alpha \norm{\mL}_F^2 \\
    \textrm{s.t.}\quad\ &\  \mL \in \setL \textrm{ and } \trace(\mL) = 2n,
\end{aligned}
\end{align}

\noindent where $\mX \in \setR^{n \times m}$ is data matrix whose columns are $\vx_i$'s and the first term quantifies the sum of the total variation of $\vx_i$'s calculated using \eqref{eq:smoothness-node-domain}. The second term is a regularizer that controls the edge density of the learned graph where higher values of the hyperparameter $\alpha$ result in a denser graph. $\mL$ is constrained to be in $\setL=\{\mL : L_{ij} = L_{ji} \leq 0\ \forall i\neq j,\ \mL \ones = \zeros\}$, which is the set of valid Laplacians. The second constraint is added to prevent the trivial solution $\mL = \zeros$. 

\section{Signed Graph Learning}
\label{sec:method}

In this section, the proposed signed graph learning method is described. We first define the GSO for signed graphs and employ the GSO to specify GFT and graph filtering operations for signed graphs. We then learn an unknown signed graph $G$ from smooth signed graph signals. A computationally efficient version of the proposed method is also developed based on large scale graph learning literature \cite{kalofolias2017large}.

\subsection{Net Laplacian For Signed GSP}
\label{ssec:method-signed-gsp}

In order to model the relationship between graph signals and a signed graph, one needs to determine the GSO, from which GFT and graph filtering are derived. Negative edges in signed graphs make defining a GSO challenging, since one needs to determine the effect of negative edges on graph signals. Recent work on signed GSP \cite{dittrich2020signal} proposes to employ the signed Laplacian \cite{kunegis2010spectral}, which is defined as $\mL_{\rm s} = \diag(|\mA|\ones) - \mA$ where $|\mA|$ denotes the absolute value of $\mA$. In particular, the diagonal entries of $\mL_{\rm s}$ are forced to be large enough to make it a positive semi-definite matrix. Although a positive graph spectrum makes interpretation of GFT easier, its total variation may not be appropriate for some applications. To this end, we propose to employ an alternative GSO, called \textit{the net Laplacian} \cite{stanic2020spectrum}, for signed GSP.

The net Laplacian of a signed graph is defined based on the standard definition of the Laplacian matrix for unsigned graphs as $\mL_{\rm n} = \diag(\mA \ones) - \mA$\footnote{Even though ``the net Laplacian" is just standard Laplacian applied to a signed graph, we prefer to use this term to emphasize the fact that we are working with signed graphs.}. Since it is a symmetric matrix, its eigendecomposition is $\mL_{\rm n} = \mV_{\rm n} \mLambda_{\rm n} \mV_{\rm n}^\top$ where the columns of $\mV_{\rm n}$ are eigenvectors and $\mLambda_{\rm n}$ is the diagonal matrix of eigenvalues $\lambda_{{\rm n}, 1} \leq \lambda_{{\rm n}, 2} \leq \dots \leq \lambda_{{\rm n}, n}$. Since the spectrum of net Laplacian is indefinite, its quadratic form can take on positive and negative values, which makes it unsuitable to define a total variation as in \eqref{eq:smoothness-node-domain}. Inspired by \cite{petrovic2020communityaware}, we shift $\mL_{\rm n}$'s spectrum to overcome this issue and introduce $\tilde{\mL} = \mL_{\rm n} + \gamma \mI$ where $\gamma \geq |\lambda_{{\rm n}, 1}|$ is a large enough constant to make sure $\tilde{\mL}$ is a positive semi-definite matrix.
We can define the total variation of a signal using $\tilde{\mL}$ as follows:
\begin{align}
    \label{eq:signed-smoothness-node-domain}
    \vx^\top \tilde{\mL} \vx = \frac{1}{2}\sum_{i\neq j}^n s_{ij} w_{ij} (x_i - x_j)^2 + \gamma \sum_{i=1}^n x_i^2,
\end{align}

\noindent where we use the fact $A_{ij} = s_{ij} w_{ij}$. Smaller values of \eqref{eq:signed-smoothness-node-domain} reflect low total variation of $\vx$ with respect to the signed graph in the sense that the difference $x_i - x_j$ is (i) small for $s_{ij} = 1$, and (ii) large for $s_{ij} = -1$. In particular, this total variation definition takes edge signs into account such that low variation across the graph requires signal values to be similar (dissimilar) at nodes connected by a positive (negative) edge where similarity of signals is defined based on their difference, i.e. $x_i - x_j$, as in \eqref{eq:smoothness-node-domain} for unsigned graphs. Using \eqref{eq:signed-smoothness-node-domain}, we can assign a notion of frequency to the spectrum of the net Laplacian. Namely, for eigenvector $\vv_{{\rm n}, i}$, we have $\vv_{{\rm n}, i}^\top \tilde{\mL} \vv_{{\rm n}, i} = \lambda_{{\rm n}, i} + \gamma$. Thus, small (large) eigenvalues of the net Laplacian correspond to low (high)-frequencies. With this notion of frequency, GFT and graph filtering for signed graphs can be defined in the same way as unsigned graphs. Namely, signed GFT is given by $\widehat{\vx} = \mV_{{\rm n}}^\top \vx$ and inverse signed GFT is $\vx = \mV_{{\rm n}} \widehat{\vx}$. A signed graph filter can be defined as $\mH = \mV_{{\rm n}} h(\mLambda_{{\rm n}}) \mV_{{\rm n}}^\top$ where $h(\cdot)$ is the frequency response of the signed graph filter. 

\begin{remark}
    \label{remark:netl-vs-sl}
    The signed Laplacian $\mL_{\rm s}$ has recently been employed as a GSO for  signed graphs \cite{dittrich2020signal}. Signed Laplacian based total variation of $\vx$ is:
    \begin{align}
        \label{eq:signed-laplacian-smoothness-node-domain}
        \vx^\top \mL_{\rm s} \vx = \frac{1}{2}\sum_{i \neq j} w_{ij} (x_{i} - s_{ij} x_{i})^2,
    \end{align}
    
    \noindent whose value is small ``if the signal values across the positive edges are similar and the signal values across the negative edges are dissimilar" \cite{dittrich2020signal} where similarity is quantified by $x_i - x_j$ for positive edges, while dissimilarity is quantified by $x_i + x_j$ for negative edges. Thus, the difference between the signed Laplacian and net Laplacian based total variation is the way the dissimilarity is defined across negative edges, i.e., $x_i + x_j$ versus $x_i - x_j$. In the former, the dissimilarity implies $x_i \approx - x_j$, i.e., signals at node pairs connected with negative edges have opposite signs. However, this can be problematic in applications where all signals have the same signs. For example, gene expression data is all non-negative as it measures gene counts in cells or tissues \cite{chen2019single}. On the other hand, \eqref{eq:signed-smoothness-node-domain} does not have this issue. 
\end{remark}

\subsection{Signed Graph Learning from Smooth Signed Graph Signals}
\label{ssec:method-signed-gl}

Given $m$ graph signals $\{\vx_i\}_{i=1}^m$ that are defined on an unknown signed graph $G$, our goal is to learn $G$ from the given signals based on the assumption that $\vx_i$'s are the outputs of a low-pass signed graph filter. As in the case of unsigned graphs, by modeling the relationship between graph signals and $G$ through a low-pass signed graph filter, we assume $\vx_i$'s have low total variation over $G$. Thus, the unknown graph can be learned by minimizing the quadratic form of the net Laplacian with respect to the net Laplacian matrix. In order to formulate the optimization problem, we note that the net Laplacian $\mL_{\rm n}$ can be decomposed as $\mL_{\rm n} = \mL^+ - \mL^-$ where $\mL^+ = \diag(\mA^+ \ones) - \mA^+$ and $\mL^- = \diag(\mA^- \ones) - \mA^-$, which are the Laplacian matrices of the unsigned graphs constructed from the positive and negative edges, respectively. Letting $\mX \in \setR^{n \times m}$ be the data matrix whose columns are $\vx_i$'s, we then propose the following optimization problem to learn $G$: 
\begin{align}
    \minimize_{\mL^+, \mL^-} &\ \trace(\mX^\top[\mL^+ - \mL^-]\mX) + \alpha_1 \norm{\mL^+}_F^2 + \alpha_2 \norm{\mL^-}_F^2 \notag \\ 
    \textrm{s.t.}\quad\ &\ \trace(\mL^+) = 2n; \trace(\mL^-) = 2n;  \mL^+, \mL^- \in \setL; \label{eq:signed-gl} \\ 
    &\ (\mL^+, \mL^-) \in \setC, \notag
\end{align}

\noindent where the first term measures the total variation of the graph signals and is equal to the first term of \eqref{eq:signed-smoothness-node-domain}\footnote{Second term of \eqref{eq:signed-smoothness-node-domain} can be omitted in our optimization, since it does not depend on $\mL^+$ and $\mL^-$ for a fixed large enough $\gamma$.}. Frobenius norm regularization terms control the edge density of the learned $G$ such that larger $\alpha_1$ and $\alpha_2$ values lead to a signed graph with larger number of positive and negative edges. The first three constraints are the same as the ones used in unsigned graph learning given in \eqref{eq:unsigned-gl}. The last constraint restricts $\mL^+$ and $\mL^-$ to be in the set $\setC = \{(\mA, \mB)|\mA \in \setR^{n \times n}, \mB \in \setR^{n \times n}, A_{ij}B_{ij} = 0\ \forall i \neq j\}$. This constraint ensures that off-diagonal entries of $\mL^+$ and $\mL^-$ cannot be non-zero for the same element, since an edge in a signed graph is either positive or negative. 

\subsection{Optimization}
\label{ssec:method-optimization}
In order to solve the problem in \eqref{eq:signed-gl}, we vectorize the matrices such that strictly upper triangular parts for $\mL^+$ and $\mL^-$ are learned. Let $\upper(\cdot)$ be an operator that takes a symmetric matrix $\mX \in \setR^{n\times n}$ as input and returns its strictly upper triangular part as a vector $\vx \in \setR^{n(n-1)/2}$ constructed in row-major order. Also let $\mQ \in \setR^{n \times n(n-1)/2}$ be a matrix such that $\mQ \upper(\mX) = \mX \ones - \diag(\mX)$ where $\mX$ is a symmetric matrix. Define the following vectors: $\vk = 2\upper(\mX\mX^\top) - \mQ^\top \diag(\mX\mX^\top)$, $\vell^+ = \upper(\mL^+)$, and $\vell^- = \upper(\mL^-)$. The vectorized form of \eqref{eq:signed-gl} is then: 
\begin{align}%
\label{eq:vectorized-signed-gl}
\begin{split}    
    \minimize_{\vell^+, \vell^-} &\ \vk^\top\vell^+ - \vk^\top \vell^- + \alpha_1 \norm{\vell^+}_{\mP}^2 + \alpha_2 \norm{\vell^-}_{\mP}^2 \\ 
    \textrm{s.t.}\quad\ &\ \ones^\top\vell^+ = -n; \ones^\top\vell^- = -n; \\ 
    &\ \vell^+ \leq \zeros, \vell^- \leq \zeros; 
    (\vell^+)^\top \vell^- = 0,
\end{split}
\end{align}

\noindent where the first two terms correspond to the total variation term in \eqref{eq:signed-gl}. The remaining two terms measure the Frobenius norms of $\mL^+$ and $\mL^-$ where $\norm{\cdot}_\mP$ is a vector norm induced by the positive definite matrix $\mP = \mQ^\top \mQ + 2\mI$, i.e., $\norm{\vv}_\mP^2 = \vv^\top \mP \vv$ for a vector $\vv$ of appropriate dimensions. The first two constraints are equal to trace constraints in \eqref{eq:signed-gl} and the remaining constraints ensure that the learned Laplacians are in $\setL$ and $\setC$. 

The vectorized problem can be solved using Alternating Direction Method of Multipliers (ADMM). Let the objective function in \eqref{eq:vectorized-signed-gl} be represented as $f(\vell^+, \vell^-)$. Define a set $\calH_1 = \{\vv \in \setR^{n(n-1)/2} | \ones^\top \vv = -n\}$ and let $\imath_1(\cdot)$ be the indicator function whose value is $0$ if its input vector is in $\calH_1$, and $\infty$, otherwise. Similarly, define the set $\calH_2 = \{(\vv, \vw) | \vv, \vw \in \setR^{n(n-1)/2}, \vv \leq \zeros, \vw \leq \zeros, \vv^\top \vw = 0 \}$ and its indicator function $\imath_2(\cdot, \cdot)$. We can rewrite the problem in \eqref{eq:vectorized-signed-gl} by introducing two slack variables $\vz^+ = \vell^+$ and $\vz^- = \vell^-$:
\begin{align}
\label{eq:vectorized-signed-gl-admm-form}
\begin{split}    
    \minimize_{\vell^+, \vell^-, \vz^+, \vz^-} &\ f(\vell^+, \vell^-) + \imath_1(\vell^+) + \imath_1(\vell^-) + \imath_2(\vz^+, \vz^-) \\ 
    \textrm{s.t.}\quad\ &\ \vz^+ = \vell^+, \vz^- = \vell^-,
\end{split}
\end{align}

\noindent whose ADMM steps at the $k$th iteration are given as follows:
\begin{align}    
{\vz^+}^{(k+1)}, {\vz^-}^{(k+1)} & = \argmin_{\vz^+, \vz^-}\ 
    \imath_2(\vz^+, \vz^-) \notag \\  
    & + 
    \norm{\vz^+ - {\vell^+}^{(k)} + \frac{1}{\rho}{\vy^+}^{(k)}} \notag \\
    & + 
    \norm{{\vz^-} - {\vell^-}^{(k)} + \frac{1}{\rho}{\vy^-}^{(k)}},
    \label{eq:admm-z-step} \\
{\vell^+}^{(k+1)}, {\vell^-}^{(k+1)} & = \argmin_{\vell^+, \vell^-} 
    f(\vell^+, \vell^-) + \imath_1(\vell^+) + \imath_1(\vell^-) \notag \\ 
    & + \frac{\rho}{2}\norm{{\vz^+}^{(k+1)} - \vell^+ + \frac{1}{\rho}{\vy^+}^{(k)}} \notag \\
    & + \frac{\rho}{2}\norm{{\vz^-}^{(k+1)} - \vell^- + \frac{1}{\rho}{\vy^-}^{(k)}},
    \label{eq:admm-l-step} \\
{\vy^+}^{(k+1)} & = {\vy^+}^{(k)} + \rho({\vz^+}^{(k+1)} - {\vell^+}^{(k+1)}), \\
{\vy^-}^{(k+1)} & = {\vy^-}^{(k)} + \rho({\vz^-}^{(k+1)} - {\vell^-}^{(k+1)}),
\end{align}

\noindent where $\vy^+$ and $\vy^-$ are the Lagrangian multipliers associated with the constraints $\vz^+ = \vell^+$ and $\vz^- = \vell^-$, respectively. $\rho > 0$ is the ADMM parameter and superscript $^{(k)}$ represents a variable's value at the $k$th iteration. The solution of subproblem in \eqref{eq:admm-z-step} is a projection onto $\calH_2$. Namely, let $\vv = {\vell^+}^{(k)} - \frac{1}{\rho}{\vy^+}^{(k)}$ and $\vw = {\vell^-}^{(k)} - \frac{1}{\rho}{\vy^-}^{(k)}$, then: %
\begin{align}
    {z_i^+}^{(k+1)} & = \begin{cases} 
        v_i & \text{ if } v_i < w_i \text{ and } v_i < 0,\\
        0 & \text{ otherwise},
    \end{cases} \label{eq:admm-z-plus-solution} \\ 
    {z_i^-}^{(k+1)} & = \begin{cases} 
        w_i & \text{ if } w_i \leq v_i \text{ and } w_i < 0,\\
        0 & \text{ otherwise}. \label{eq:admm-z-negative-solution} 
    \end{cases}
\end{align} %

The subproblem in \eqref{eq:admm-l-step} can be solved separately for $\vell^+$ and $\vell^-$. Its solution with respect to $\vell^+$ can be found from the Karush–Kuhn–Tucker (KKT) conditions as the projection of $[(4\alpha_1 + \rho) \mI + 2 \alpha_1\mQ^\top \mQ]^{-1}(\rho {\vz^+}^{(k+1)} + {\vy^+}^{(k)} - \vk)$ onto $\calH_1$. Similarly, its solution with respect to $\vell^-$ is the projection of $[(4\alpha_2 + \rho) \mI + 2 \alpha_2\mQ^\top \mQ]^{-1}(\rho {\vz^-}^{(k+1)} + {\vy^-}^{(k)} + \vk)$ onto $\calH_1$. 

The ADMM based solution of \eqref{eq:vectorized-signed-gl} is summarized in Algorithm \ref{alg:signed-gl} where $\Pi_1$ indicates the projection operator onto $\calH_1$. ADMM iterations continue until convergence is achieved or maximum number of iterations are reached. The algorithm convergence is determined by the difference between values of the objective function in \eqref{eq:vectorized-signed-gl} at consecutive iterations (Line \ref{alg-line:convergence} in Algorithm \ref{alg:signed-gl}). Note that the proposed signed graph learning problem is non-convex due to the constraints ${\vell^+}^\top \vell^- = 0$, which along with $\vell^+ \leq \zeros, \vell^- \leq \zeros$ are known as complementarity constraints \cite{scheel2000mathematical}. ADMM is shown to converge for problems with complementarity constraints under some conditions \cite{wang2019global}, which are satisfied by the proposed problem as discussed in Section \ref{ssec:opt-convergence}. 

The computational complexity of Algorithm \ref{alg:signed-gl} can be calculated from ADMM steps. The updates of $\vz^+$ and $\vz^-$ need to be done for each of their elements, which indicates there are $n(n-1)/2$ operations in this step. For $\vell^+$ and $\vell^-$, the inverse matrix has a closed form solution that can be found using matrix inversion lemma. This matrix is sparse and has $\calO(n^2)$ entries, implying $\vell^+$ and $\vell^-$ steps can be performed in $\calO(n^2)$ time. Finally, updates of Lagrangian multipliers require $n(n-1)/2$ operations. Let $I$ be the number of ADMM iterations needed for convergence, then Algorithm \ref{alg:signed-gl} has computational complexity of $\calO(In^2)$. 

\begin{figure}[t]%
\vspace{-0.5em}%
\begin{algorithm}[H]%
\caption{Signed Graph Learning} \label{alg:signed-gl}%
\begin{algorithmic}[1]
\Require $\mX, \alpha_1, \alpha_2, \rho, \textrm{MaxIter}$, $\epsilon = \num{1e-6}{}$
\Ensure $\vell^+, \vell^-$
\State $\vk \gets \upper(\mX\mX^\top) - \mQ^\top \diag(\mX\mX^\top)$
\LineComment{0em}{Initialize variables}
\State $\vell^+ \gets \zeros,\ \vell^- \gets \zeros$
\State $\vy^+ \gets \zeros,\ \vy^- \gets \zeros$
\For{$k$ in $\{1, \dots, \textrm{MaxIter}\}$ }
\State Update $\vz^+$ and $\vz^-$ with \eqref{eq:admm-z-plus-solution} and \eqref{eq:admm-z-negative-solution}
\State $\vell^+ \gets \Pi_1([(4\alpha_1 + \rho) \mI + 2 \alpha_1\mQ^\top \mQ]^{-1}(\rho {\vz^+} + {\vy^+} - \vk))$
\State $\vell^- \gets \Pi_1([(4\alpha_2 + \rho) \mI + 2 \alpha_2\mQ^\top \mQ]^{-1}(\rho {\vz^-} + {\vy^-} - \vk))$
\State ${\vy^+} \gets {\vy^+} + \rho({\vz^+} - {\vell^+})$
\State ${\vy^-} \gets {\vy^-} + \rho({\vz^-} - {\vell^-})$
\LineComment{1.5em}{Check convergence}
\IfThen{$|\textrm{Objective}[k] - \textrm{Objective}[k-1]| \leq \epsilon$}{break} \label{alg-line:convergence}
\EndFor
\end{algorithmic}%
\end{algorithm}%
\vspace{-2em}%
\end{figure}%

\subsection{Large Scale Signed Graph Learning}
\label{ssec:method-fast-signed-gl}

The quadratic computational complexity of each ADMM iteration in Algorithm \ref{alg:signed-gl}  can be prohibitive in large-scale applications where the number of nodes $n$ is large. In this section, we propose an approach to speed up Algorithm \ref{alg:signed-gl} based on large scale unsigned graph learning literature \cite{kalofolias2018large} and approximate nearest-neighbors methods \cite{dong2011efficient, malkov2018efficient}. The main idea is to reduce the number of unknown variables while ensuring the accuracy of graph learning. In particular, we can reduce the dimensions of the variables in ADMM steps from $n(n-1)/2$ to $\calO(nk)$ where $k \ll n$ is a parameter to be determined. This reduction yields an improvement in time complexity such that each ADMM iteration can be performed in $\calO(nk)$ time rather then $\calO(n^2)$. 

Consider the vectorized version of the proposed signed graph learning algorithm given in $\eqref{eq:vectorized-signed-gl}$. Assume that it is known \textit{a priori} that only some of the node pairs can make connections in the signed graph, i.e., there is a set $E_{\rm a} = \{(i, j)\ |\ i, j \in V, i < j \}$ such that any node pair not in $E_{\rm a}$ is known to be unconnected in the signed graph $G$. This set introduces a constraint for problem \eqref{eq:vectorized-signed-gl}. Introduce an index function $\calI : V \times V \to \{1, \dots, n(n-1)/2\}$ such that $\calI(i, j)$ is the index of $\vell^+$'s entry (or $\vell^-$) corresponding to $L_{ij}^+$ (or $L_{ij}^-$), i.e, $\ell_{\calI(i, j)}^+ = L_{ij}^+$ (similarly for $\mL^-$). Then, $\eqref{eq:vectorized-signed-gl}$ needs to be constrained such that $\ell_{\calI(i,j)}^+ = \ell_{\calI(i,j)}^- = 0$ if $(i,j) \notin E_{\rm a}$. The first term in the objective function \eqref{eq:vectorized-signed-gl} can then be rewritten:
\begin{align}
    \vk^\top \vell^+ = \sum_{(i, j) \in E_{\rm a}} k_{\calI(i,j)} \ell_{\calI(i,j)}^+ + \sum_{(i, j) \notin E_{\rm a}} k_{\calI(i,j)} \ell_{\calI(i,j)}^+, 
\end{align}

\noindent where the second term is $0$ since $\ell_{\calI(i,j)}^+ =0$ for $(i,j) \notin E_{\rm a}$. Let $\tilde{\vell}^+ \in \setR^{|E_{\rm a}|}$ and $\tilde{\vk} \in \setR^{|E_{\rm a}|}$ be vectors constructed by sub-sampling $\vell^+$ and $\vk$ to the entries that are in $E_{\rm a}$. Then, we have $\vk^\top \vell^+ = \tilde{\vk}^\top \tilde{\vell}^+$. The remaining terms in \eqref{eq:vectorized-signed-gl} can similarly be rewritten by introducing $\tilde{\vell}^- \in \setR^{|E_{\rm a}|}$ and $\tilde{\mP} = \tilde{\mQ}^\top \tilde{\mQ} + \mI \in \setR^{|E_{\rm a}|\times |E_{\rm a}|}$ where $\tilde{\mQ} \in \setR^{n \times |E_{\rm a}|}$ is the submatrix of $\mQ$ induced by columns corresponding to edges in $E_{\rm a}$. Thus, the following optimization problem is obtained:
\begin{align}
\label{eq:vectorized-large-scale-signed-gl}
\begin{split}    
    \minimize_{\tilde{\vell}^+, \tilde{\vell}^-} &\ \tilde{\vk}^\top \tilde{\vell}^+ - \tilde{\vk}^\top \tilde{\vell}^- + \alpha_1 \norm{\tilde{\vell}^+}_{\tilde{\mP}}^2 + \alpha_2 \norm{\tilde{\vell}^-}_{\tilde{\mP}}^2 \\ 
    \textrm{s.t.}\quad\ &\ \ones^\top \tilde{\vell}^+ = -n; \ones^\top \tilde{\vell}^- = -n; \\ 
    &\ \tilde{\vell}^+ \leq \zeros, \tilde{\vell}^- \leq \zeros; 
    (\tilde{\vell}^{+})^{\top} \tilde{\vell}^- = 0,
\end{split}
\end{align}

\noindent which can be solved with the same ADMM approach described in Section \ref{ssec:method-optimization} where each ADMM iteration has computational complexity of $\calO(|E_{\rm a}|)$ thanks to the reduction in number of variables to be learned. 

Based on the above discussion, the time complexity of Algorithm \ref{alg:signed-gl} can be reduced if one knows $E_{\rm a}$. In most applications, this set is not available, however, an approximate $E_{\rm a}$ can be constructed from the observed signals $\mX$. Following the approach proposed in \cite{kalofolias2018large} for unsigned graph learning, for each node $i$, we define neighbors $\calN_i^+$ and $\calN_i^-$ that include the nearest and farthest $k$ neighbors of node $i$, respectively. These neighbors are found by calculating the distance between $\mX_{i\cdot}$ and the remaining rows\footnote{To find the nearest and farthest neighbors, we can employ approximate nearest neighbors algorithms and their variants for farthest neighbors  \cite{dong2011efficient, malkov2018efficient, hjaltason1999distance}. These approaches find the desired neighbors in $\calO(n\log(n))$, which is more efficient than the brute-force approach of calculating the distance for all row pairs $\mX_{i\cdot}$ and $\mX_{j\cdot}$.}. $E_{\rm a}$ is constructed as $E_{\rm a} = \bigcup_{i=1}^n \{(i, j)\ |\ j \in \calN_i^+ \cup \calN_i^-\}$. The main assumption behind this choice for $E_{\rm a}$ is based on \eqref{eq:signed-smoothness-node-domain}. In particular, small values of net Laplacian quadratic form imply nodes connected by a positive edge are similar, i.e., $\norm{\mX_{i\cdot} - \mX_{j\cdot}}_2^2$ is small. Thus, the search space for positive edges in \eqref{eq:vectorized-large-scale-signed-gl} is restricted to those node pairs that are in close proximity. At the same time, negatively connected node pairs are dissimilar, i.e., $\norm{\mX_{i\cdot} - \mX_{j\cdot}}_2^2$ is large. Then, the search space for negative edges in \eqref{eq:vectorized-large-scale-signed-gl} is limited to the node pairs that are far away from each other. 

When constructing $E_{\rm a}$, $k$ determines the maximum number of positive and negative edges a node can have in the learned signed graph. Assume that we are trying to learn a signed graph with a pre-determined positive and negative edge density
$\delta^+$ and $\delta^-$. Average positive and negative node degrees for this graph are respectively $k_0^+ = \delta^+/n$ and $k_0^- = \delta^-/n$. We set $k = \lceil\beta\max(k_0^+, k_0^-)\rceil$ where $\beta > 1$ is a multiplicative factor to ensure $E_{\rm a}$ is large enough to infer the correct edges. With this way of selecting $E_{\rm a}$, computational complexity of Algorithm \ref{alg:signed-gl} is $\calO(nk)$ per ADMM iteration. Note that, since most real world networks are sparse, we have $k \ll n$, indicating lower time requirements compared to $O(n^2)$. 

\section{Theoretical Analysis}
\label{sec:theory}
\subsection{Convergence of the Algorithm}
\label{ssec:opt-convergence}

The optimization problem in (\ref{eq:vectorized-signed-gl-admm-form}) is non-convex due to the set of complimentarity constraints. The convergence of the algorithm can be shown using Theorem 1 presented in \cite{wang2019global} by proving the objective function and constraints of \eqref{eq:vectorized-signed-gl-admm-form} satisfy the assumptions A$1$-A$5$ in \cite{wang2019global}. These assumptions are satisfied by the proposed optimization problem, as stated by the following theorem whose proof can be found in Appendix \ref{appdx:optimization-convergence}:
\begin{theorem}
\label{thrm:convergence}
    For sufficiently large $\rho$, the sequence $(\vell^{(k)},\vz^{(k)},\vy^{(k)})$ obtained by Algorithm \ref{alg:signed-gl} has limit points and these limit points are stationary points of the augmented Lagrangian of \eqref{eq:vectorized-signed-gl-admm-form}.
\end{theorem}

\subsection{Estimation Error Bound}
In this section, we perform the theoretical analysis of \eqref{eq:signed-gl} to study the effect of sample size, number of nodes and the topology of the true graph structure on the estimation error bound. Since the problem is non-convex, we consider local optima, $\widehat{\mL}^+$ and $\widehat{\mL}^-$, and derive the error bound of the local optimum with respect to true Laplacian matrices $\mL_0^+$ and $\mL_0^-$. Let $\widehat{\mL}$ be a block diagonal matrix constructed from $\widehat{\mL}^+$ and $\widehat{\mL}^-$. Similarly, define the block diagonal matrix $\mL_0$. To derive the estimation error bound between $\widehat{\mL}$ and $\mL_0$, we make the following assumptions:
\begin{enumerate}
    \item[(A1)] The optimization is constrained to a local neighborhood of the true parameter matrix $\mathbf{L}_{0}$. \label{A1}
    \item[(A2)] The set of signals \(\left\{\mathbf{x}_{i}\right\}_{i=1}^{m}\) is assumed to be distributed according to a sub-Gaussian distribution where each $\vx_i$ has a mean of \(\boldsymbol{0}\) and a covariance matrix \(\mSigma_0\). \label{A2}
    \item[(A3)] For the true Laplacians $\mathbf{L}_{0}^{+}$ and $\mathbf{L}_{0}^{-}$, $\operatorname{tr}(\mathbf{L}_{0}^{+}) = \operatorname{tr}(\mathbf{L}_{0}^{-}) = 2n$. This implies that the true Laplacians satisfy $\operatorname{max}\{\left\|\mathbf{L}_{0}^{+}\right\|_{F}, \left\|\mathbf{L}_{0}^{-}\right\|_{F}\} \leq M $ for some constant $M>0$. Then, it should also be true that $\mathbf{L}_0$ and the corresponding true covariance matrix $\boldsymbol{\Sigma}_{0}$ have bounded Frobenius norms. \label{A3} 
\end{enumerate}
    
\begin{theorem}
\label{thrm:consistency}
    Under the assumptions (A1), (A2), (A3)  with regularization parameters $\alpha_{1},\alpha_{2}>0$, the estimation error is upper bounded with probability $1-2e^{-cn}$ as follows:
    \begin{align}
        \norm{\widehat{\mathbf{L}} - \mathbf{L}_{0}}_{F} \leq C_{1} \frac{n}{\alpha_{m}\sqrt{m}} + \frac{1}{\alpha_{m}}\left(\left\|\mSigma_0\right\|_{F} + 2 \tilde{\alpha}_{m}M  \right),
    \end{align}
 for $m > C_1^2 n $\ where $C_1=C+1$ and $C=C_{K}, c=c_{K}$ are positive constants depending on the sub-Gaussian norm $K= \max_{i} \left\|\mathbf{x}_{i}\right\|_{\psi_2}$ and $\alpha_{m} = \min\left\{\alpha_{1}/m,\alpha_{2}/m\right\}$, $\tilde{\alpha}_{m}= \max\left\{\alpha_{1}/m,\alpha_{2}/m\right\}$.
\end{theorem}

The proof of this theorem is given in Appendix \ref{appdx:consistency-proof}. The estimation error bound shows that the accuracy of the estimated \(\widehat{\mathbf{L}}\) is influenced by the number of nodes \( n \), the number of samples \( m \), and the choice of regularization parameters. To analyze the asymptotic behavior of the estimation error, we consider the upper bound derived in Theorem~\ref{thrm:consistency}, which decomposes into two principal terms, the sample-dependent term \( \frac{n}{\alpha_m \sqrt{m}} \), and the regularization-driven term \( \frac{1}{\alpha_m} \left( \| \boldsymbol{\Sigma}_0 \|_F + 2 \tilde{\alpha}_m M \right) \) similar to \cite{10387579,zhang2025graphlearningdatasilos}. For the overall estimation error \( \norm{\widehat{\mathbf{L}} - \mathbf{L}_0}_F \) to vanish as \( m \to \infty \), it is sufficient that \( \alpha_m = o ( \frac{n}{\sqrt{m}} ) \), and simultaneously \( \alpha_m \to \infty \) with \( \tilde{\alpha}_m \asymp \alpha_m  \). This ensures that both terms of the bound decay to zero. Under these conditions, minimum estimation error bound can be achieved in the Frobenius norm provided that \( n = o(\sqrt{m} \cdot \alpha_m) \). However, in practical scenarios, it is common to consider fixed regularization parameters where \( \alpha_m = \mathcal{O}(1) \) and \( \tilde{\alpha}_m = \mathcal{O}(1) \). In such cases, the estimation error bound simplifies to
\[
\left\| \widehat{\mathbf{L}} - \mathbf{L}_0 \right\|_F \leq C_1 \frac{n}{\sqrt{m}} + C_2,
\]
where \( C_2 = \| \boldsymbol{\Sigma}_0 \|_F + 2M \) represents an irreducible error  determined by the underlying covariance structure and the bound of the norm of the true Laplacians. Consequently, the error does not converge to zero even as \( m \to \infty \).

\section{Results}
\label{sec:results}
In this section, we evaluate the performance of our proposed signed graph learning approaches and benchmark them against existing methods under different setups. We use SGL and fastSGL to refer Algorithm 1 and its large-scale version, respectively. 

\subsection{Simulated Datasets}

The proposed algorithms are first evaluated on synthetic datasets in which the true signed graph structure is known.  We compare with two existing signed graph learning methods to benchmark the proposed methods. We consider the method developed in \cite{matz2020learning}, where the smoothness of the observed data is measured using the signed Laplacian \cite{kunegis2010spectral} as GSO. The unknown signed graph is learned by minimizing smoothness with respect to the signed Laplacian through a two-step optimization process. The first step learns the signs of the edges and the second step employs unsigned graph learning to infer the topology. For the second step, we solve \eqref{eq:unsigned-gl} with ADMM and refer to this method as signed Laplacian learning (SLL). The second method \cite{fong2024efficient} is based on our previous work \cite{karaaslanli2022scsgl}; however, it learns the adjacency matrix rather than the net Laplacian. The modified optimization problem is solved using proximal ADMM (pADMM) \cite{fong2024efficient} and thus we refer to this method as pADMM. 

\noindent \textit{Data Generation:} In order to generate simulated data, we first construct a signed graph $G$ with $n$ nodes from three random graph models: signed versions of Erdős–Rényi (ER), Barabási–Albert (BA) and random geometric graph (RGG). For ER and BA models, we employ the process described in \cite{yokota2025efficient} to obtain the signed graphs. First, an unsigned graph from ER model with edge probability parameter $p$ or BA model with growth parameter $m_{\rm BA}$ is constructed. The signed graph, $G$, is then constructed from the unsigned graph by assigning signs to edges using balance theory \cite{kirkley2019balance}. In particular, each node $i$ is assigned to a polarity $s_i$, which is $+1$ with probability $0.5$ and $-1$, otherwise. Edges that connect node pairs that have the same polarity are set as positive, while any edge connecting two nodes with different polarities are set as negative. This process generates a balanced signed graph. Since real world graphs can have a degree of imbalance, we flip the signs of $\zeta$ fraction of edges to introduce a level of imbalance in $G$. In the RGG model, $G$ is generated by first drawing $n$ points from 2D unit square where each point represents a node in $G$. Nodes are then connected to their $k_{\rm RGG}$ nearest and farthest neighbors with positive and negative edges, respectively. 

Given the graph $G$, we generate $m$ graph signals $\{\vx_i\}_{i=1}^m$. Each signal is $\vx_i = \mH\vx_0$ where $\mathbf{x}_0 \sim \calN(\zeros, \mI)$ and $\mH$ is a low-pass signed graph filter. We use three different filters: Gaussian, Heat and Tikhonov filters \cite{kalofolias2016learn}. Each filter is constructed as $\mH = \mV_{\rm n} h(\widehat{\mLambda}_{\rm n})\mV_{\rm n}^\top$ where $\widehat{\mLambda}_{\rm n}$ is the diagonal matrix of eigenvalues of $\mL_{\rm n}$ shifted and scaled to $[0,1]$. Filters are defined as $h(\widehat{\mLambda}_{\rm n}) = (\widehat{\mLambda}_{\rm n} + \eta \mI)^{-1}$ (Gaussian), $h(\widehat{\mLambda}_{\rm n}) = \exp(-\eta \widehat{\mLambda}_{\rm n})$ (Heat), and $h(\widehat{\mLambda}_{\rm n}) = (\mI + \eta \widehat{\mLambda}_{\rm n})^{-1}$ (Tikhonov) where $\eta$ is the filter parameter. We finally add $\varepsilon \%$ noise (in $\ell_2$-norm sense) to each generated $\mathbf{x}_{i}$. We repeat each simulation $20$ times with different seeds and report the average performance. 

\noindent \textit{Performance Metric:} Signed graph learning can be considered as a classification problem, where each edge belongs to one of three classes: +1 for positive edges, -1 for negative edges and 0 for unconnected node pairs. Therefore, to evaluate the performance, F1 score with macro averaging is utilized. In particular, we first construct an adjacency matrix $\widehat{\mA}$, where $\widehat{A}_{ij} = +1$ if a method finds a positive edge between nodes $i$ and $j$; $\widehat{A}_{ij}=-1$ if a method finds a negative edge; and $\widehat{A}_{ij} = 0$, otherwise. Given the ground truth adjacency matrix $\mA$, macro F1 is equal to $[\textrm{F1}(\widehat{\mA}^+, \mA^+) + \textrm{F1}(\widehat{\mA}^-, \mA^-)]/2$.

\noindent \textit{Hyperparameter Selection:} To evaluate the performance of the proposed methods SGL, fastSGL against SLL and pADMM, proper hyperparameter selection is important to determine the weights of the different terms within the objective functions. For both SGL and fastSGL, the hyperparameters $\alpha_{1}, \alpha_{2}$ determine the density of the learned graph. SLL also has the same type of regularization parameter $\mu$ which controls the density of the graph. The objective function of pADMM has two hyperparameters of interest: $\alpha$, and $\beta$, which control the sparsity. A grid search is performed over the different appropriate ranges of hyperparameters for all methods, and we select the configuration that yields the highest macro F1 score. Finally, the hyperparameter $k$ in fastSGL, which determines the maximum number of positive and negative edges a node can have, is set to $20$ in all experiments without performing any grid search. 

\noindent \textit{Experiment 1 - Comparison across graph models and filters:} In the first experiment, we evaluate how different methods perform for different signed graph models and filters. We generate $G$ with $n=100$ nodes from one of the three random graph models: signed ER model with $p=0.1$ and $\zeta=0.1$; signed BA model with $m_{\rm BA}=5$ and $\zeta=0.1$ and RGG with $k_{\rm RGG}=5$. $m=2000$ signals with $\varepsilon=10\%$ noise are then generated using one of the three filters, with parameters $\eta = 0.1$, $\eta = 2$ and $\eta = 5$ for Gaussian, Heat and Tikhonov, respectively. The methods are applied to the generated data where the range of grid search is: $\alpha_{1},\alpha_{2}\in \{10^{-3 + r/10}\ |\ r = 0,\dots,20\}$; $\{10^{-3 + r/25}\ |\ r = 0,\dots,50\}$ for $\mu$; and $\{10^{-2 + r/10}\ |\ r = 0,\dots,20\}$ for $\alpha$ and $\beta$. 

\begin{figure}[t]
    \centering
    \includegraphics[width=\linewidth]{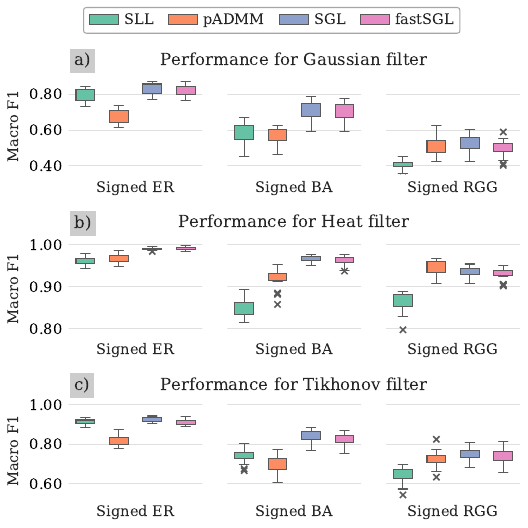}
    \caption{Performance of methods across different random graph models and filters. Macro F1 scores are reported when graph signals are generated using (a) Gaussian filter ($\eta = 0.1$), (b) Heat filter ($\eta = 2$) (b), and (c) Tikhonov filter ($\eta = 5$). Graphs are generated with signed versions of ER model ($p=0.1, \zeta=0.1$), BA model ($m_{\rm BA}=5, \zeta=0.1$) and RGG model ($k_{\rm RGG}=5$).}
    \label{fig:comparison}
\end{figure}

In Figure \ref{fig:comparison}, we report Macro F1 scores for all nine combinations of signed graph models and filters. With the exception of few cases, the proposed SGL method obtains the highest Macro F1 score. fastSGL has slightly worse performance, which is expected since there are potentially some true edges not included in the allowed edge set $E_{\rm a}$. However, it performs well with lower time complexity as will be shown in the next experiment. Existing methods show higher variability in performance across different graph models. For example, SLL performs well for signed ER graphs and while its performance drops for the other two graph models. Similarly, pADMM has high performance for the RGG model, while it performs poorly for the other two. On the other hand, SGL provides good results across different graph models and filters. 

\noindent \textit{Experiment 2 - Time complexity analysis:} In this experiment, we compare the different methods based on their time complexity and investigate if fastSGL can provide the desired reduction in time requirements. We construct $G$ using the signed BA model with $m_{\rm BA}=5$, $\zeta = 0.1$ and signals are generated using the Heat filter with $\eta = 2$. We vary the number of nodes $n$ from $200$ to $1000$ with step size of $200$. For each $n$, we generate $m=20*n$ signals with $\varepsilon=10\%$ noise. By increasing $m$ with $n$, our goal is to reduce the effect of sample size on time complexity, since our aim is to observe how time complexity of methods is influenced by $n$. Ranges for grid search are set as $\{10^{-3 + r/10}\ |\ r = 0,\dots,10\}$ for $\alpha_1$ and $\alpha_2$; $\{10^{-3 + r/25}\ |\ r = 0,\dots,50\}$ for $\mu$; and $\{10^{-2.5 + r/6.6}\ |\ r = 0,\dots,10\}$ for $\alpha$ and $\beta$. In particular, ranges for $\alpha_1$, $\alpha_2$, $\alpha$ and $\beta$ are reduced  due to increased computational requirements to conduct the experiment. 

\begin{figure}[t]
    \centering
    \includegraphics[width=\linewidth]{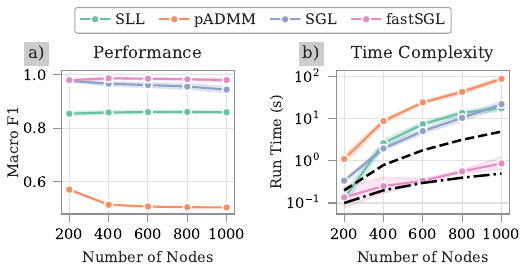}
    \caption{Performance and time complexity analysis of methods with increasing number of nodes. (a) plots the Macro F1 scores as a function of number of nodes. (b) reports time (in seconds) required to run each algorithm. Black dashed and dotted dashed lines are the quadratic and linear functions of number of nodes, respectively.}
    \label{fig:time-complexity}
\end{figure}

Figure \ref{fig:time-complexity} shows both the macro F1 scores and run times as a function of $n$. The performances of SGL, fastSGL and SLL are mostly stable with increasing number of nodes and are inline with the first experiment. There is a slight drop in SGL performance when $n$ increases, but we believe this is due to the reduced range of grid search. pADMM performs poorly as the learned graphs have very high edge density, which leads to its low performance.  In terms of time complexity,  fastSGL has the lowest run time as expected. Figure \ref{fig:time-complexity}b also shows quadratic and linear functions of $n$ as  dashed and dotted dashed lines, respectively. It can be seen that run times of SLL, SGL and pADMM follow the quadratic function whereas the run time of fastSGL follows the linear function as $|E_{\rm a}|=40n$.

\noindent \textit{Experiment 3 - Parameter sensitivity:}
For the last experiment, the effect of different data generation parameters on the performance is studied. Simulated data is generated with varying number of signals $m$; imbalance $\zeta$ of the graph; and noise level $\varepsilon$. Signed BA model with $m=5$ is employed to generate a graph with $n=100$ nodes and Heat filter with parameter $\eta = 2$ is used for signal generation. The range of grid search is set to be the same as Experiment 1. 

\begin{figure*}[t]
    \centering
    \includegraphics[width=\linewidth]{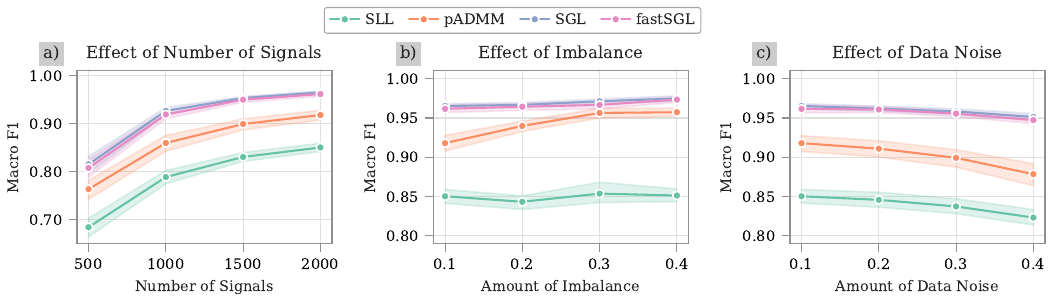}
    \caption{Performance analysis of methods with respect to different parameters of the data generation process. Macro F1 scores are reported as a function of (a) number of signals, (b) amount of imbalance in the signed graph, (c) amount of noise in graph signals. Shaded areas are confidence intervals calculated over 20 repetitions.}
    \label{fig:parameter-sensitivity}
\end{figure*}

In Figure \ref{fig:parameter-sensitivity}a, Macro F1 score as a function of the number of signals is reported. As expected, for all methods larger sample size leads to higher performance. The effect of balance in the ground truth signed graph is shown in Figure \ref{fig:parameter-sensitivity}b for all methods. Since none of the methods impose any constraints to ensure that the learned signed graphs are balanced, Macro F1 scores of all methods except pADMM remain the same. Interestingly, there is a slight increase in pADMM's performance when we increase the imbalance of the underlying graph. Figure \ref{fig:parameter-sensitivity}c shows that  all methods' performance drops with increased noise level with SGL and fastSGL being the most robust to noise. Finally, method comparison in all three subplots of Figure \ref{fig:parameter-sensitivity} is inline with that reported in Figure \ref{fig:comparison}: i.e., SGL has the highest performance and it is followed by fastSGL, pADMM and SLL.

\subsection{Gene Regulatory Network Inference}

In this section, the proposed approach is tested in the gene regulatory network (GRN) inference problem based on the benchmarking framework developed in \cite{pratapa2020benchmarking}. Regulatory interactions within living cells can be modeled as a signed graph where nodes correspond to genes and positive and negative edges represent the activating and inhibitory relations between them, respectively \cite{badia2023gene}. GRNs govern the expression of genes within cells; as such, activation indicates increased co-expression of the gene pair, while inhibition results in decreased co-expression \cite{kim2023deciphering}. Thus, transcriptomics data, \textit{e.g.} single-cell RNA sequencing (scRNA-seq), which measures gene expression levels, can be used to infer GRNs where similarity and dissimilarity between expression levels of gene pairs are used to learn whether they are connected with a positive or a negative edge. 

We applied SGL, SLL and pADMM, to four curated datasets used in \cite{pratapa2020benchmarking}. Due to the small size of the datasets, we do not apply fastSGL. Each data set is generated synthetically using the BoolODE framework, which simulates scRNA-seq data from a given Boolean GRN model describing regulatory relations between genes. BoolODE employs non-linear ordinary differential equations to generate scRNA-seq data that well-simulates various features of real scRNA-seq datasets, such as dropout, cellular differentiation, and heterogeneity. Four Boolean models related to tissue development are provided as input to BoolODE to generate datasets: mammalian cortical area development (mCAD), ventral spinal cord (VSC) development, hematopoietic stem cell (HSC) differentiation, and gonadal sex determination (GSD). The statistics of the ground-truth GRNs are reported in Table \ref{tbl:grn-gt-stats}. For each Boolean model, 10 simulated datasets, each with 2000 cells, are generated, and the average performance across 10 simulations is reported. A key characteristic of scRNA-seq is dropout, where the expression level of many genes is unobserved or zero. Two additional versions of the datasets with dropout rates $50\%$ and $70\%$ are also analyzed to observe how the methods are affected by the increased number of zeros in the scRNA-seq. More details on the data sets can be found in \cite{pratapa2020benchmarking}.

\begin{table}[t]
    \caption{Curated Dataset Statistics}
    \label{tbl:grn-gt-stats}
    \centering
    \footnotesize
    \begin{tabular}{lccc}
        \toprule
        & No. of Nodes & No. of Pos. Edges & No. of Neg. Edges \\ \cmidrule{2-4} 
        GSD & 19 & 38 & 20 \\
        HSC & 11 & 9 & 11 \\ 
        VSC & 8 & 0 & 10 \\
        mCAD & 5 & 3 & 6 \\
        \bottomrule
    \end{tabular}
\end{table}

\begin{table*}[t]
    \caption{Performance of methods on four curated gene regulation datasets of \cite{pratapa2020benchmarking}. For VSC, AUPRC ratio is only calculated for negative edges, since its ground truth GRN does not include positive connections. No negative edges are identified by SLL for VSC datasets, therefore the performance are not reported for that case.}
    \label{tbl:grn-performance}
    \centering
    \footnotesize
    \begin{tabular}{l@{\hspace{3em}}ccc@{\hspace{4em}}ccc@{\hspace{4em}}ccc}
        \toprule
        & \multicolumn{3}{c@{\hspace{4em}}}{No Dropout}  & \multicolumn{3}{c@{\hspace{4em}}}{50\% Dropout} & \multicolumn{3}{c}{70\% Dropout} \\ \cmidrule(r{4em}){2-4} \cmidrule(r{4em}){5-7} \cmidrule(r){8-10}
        & SLL & pADMM & SGL & SLL & pADMM & SGL & SLL & pADMM & SGL \\ \midrule
        GSD  & 1.78 & 2.32 & {\bf 2.90} & 1.76 & 2.46 & {\bf 2.96} & 1.72 & 2.50 & {\bf 2.92} \\
        HSC  & 2.48 & 3.52 & {\bf 3.78} & 2.79 & 3.68 & {\bf 3.90} & 2.85 & 3.82 & {\bf 4.07} \\
        VSC  & $-$  & 2.29 & {\bf 2.59} & $-$  & 2.32 & {\bf 2.62} & $-$  & 2.40 & {\bf 2.62} \\
        mCAD & 1.85 & 2.18 & {\bf 2.39} & 1.88 & 2.17 & {\bf 2.39} & 2.17 & 2.18 & {\bf 2.41} \\
        \bottomrule
    \end{tabular}
\end{table*}

The accuracy of GRN inference is quantified using the area under the precision-recall curve ratio (AUPRC ratio) to be in line with \cite{pratapa2020benchmarking}. In particular, AUPRC values are calculated separately for positive and negative edges, which we indicate as $\textrm{AUPRC}^+$ and $\textrm{AUPRC}^-$. Both values are then divided by positive and negative AUPRC values of a random estimator shown as $\textrm{AUPRC}_{\rm r}^+$ and $\textrm{AUPRC}_{\rm r}^-$. AUPRC ratio is then $(\frac{\textrm{AUPRC}^+}{\textrm{AUPRC}_{\rm r}^+} + \frac{\textrm{AUPRC}^-}{\textrm{AUPRC}_{\rm r}^-})/2$, which specifies how much better than a random estimator a method performs. Note that, for a ground truth network with $\delta^+$ positive edge density and $\delta^-$ negative edge density, we have $\textrm{AUPRC}_{\rm r}^+ = \delta^+$ and $\textrm{AUPRC}_{\rm r}^- = \delta^-$. 

Table \ref{tbl:grn-performance} reports the performance of three methods across all four datasets and different dropout rates, where the highest AUPRC ratio for each case is typed in boldface. The hyperparameters of the methods are selected as in simulated data, where a grid search was performed, and the hyperparameter providing the highest AUPRC ratio is reported. Across all datasets and for all three dropout rates, the highest performance is achieved by SGL, followed by pADMM. The lowest performing method is SLL. When graphs learned by SLL are further analyzed, it is observed that they include only positive edges, which results in low performance. As discussed in Remark \ref{remark:netl-vs-sl}, when learning signed graphs, SLL defines the dissimilarity  as signals at nodes connected with negative edges having opposite signs. Since gene expression data is non-negative, SLL fails to identify negative edges. 

The effect of increasing dropouts differs across different models. SLL's AUPRC ratio increases for HSC and mCAD, while there is a drop for GSD. pADMM exhibits higher performance as the dropout rate increases for GSD, HSC, and VSC, whereas no significant change is observed for mCAD. Finally, SGL improves with increasing dropout rate for HSC and there is no clear trend for the remaining models. In most cases, changes in the AUPRC ratio with the growing number of dropouts are moderate. Considering the diverse features of Boolean models (see Table \ref{tbl:grn-gt-stats}), we conclude that these observations are due to models' network characteristics rather than a systemic influence of dropout on the methods. 

\section{Conclusions}

In this paper, we introduced a signed graph learning algorithm based on the smoothness assumption for the observed signals. In particular, we formulated the problem by minimizing the total variation with respect to the net Laplacian and wrote the learning problem in terms of the positive and negative graph Laplacians. Unlike the signed Laplacian definition, this approach allows a more flexible definition of similarity/dissimilarity of signals with respect to the underlying graph. We introduced two algorithms, i.e., exact and fast (approximate), to solve the resulting optimization problem and evaluated the performance for different graph structures and signals with respect to existing signed graph learning algorithms. The performance of SGL on both simulated graph signals and scRNA-seq data illustrate the robustness and flexibility of our method for different noise levels, graph structures and signal generation parameters.

\appendices 

\section{Proof of Theorem 1}
\label{appdx:optimization-convergence}

The proof of Theorem \ref{thrm:convergence} follows from showing that the optimization problem in \eqref{eq:vectorized-signed-gl-admm-form} satisfies the assumptions A$1$-A$5$ given in \cite{wang2019global}. To prove that assumption A$1$ is satisfied, we need to show that the objective function in (\ref{eq:vectorized-signed-gl-admm-form}), i.e., $f(\vell^+, \vell^-) + \imath_1(\vell^+) + \imath_1(\vell^-) + \imath_2(\vz^+, \vz^-)$, is coercive over the constraint set $\mathcal{F}:=\left\{\left(\vz^{+}, \vz^{-}, \vell^{+}, \vell^{-}\right): \vz^{+}-\vell^{+}=\mathbf{0}, \vz^{-}-\vell^{-}=\mathbf{0}\right\}$. The first term of the objective function is defined as:
\begin{align*}
    f(\vell^+, \vell^-) = \vk^\top (\vell^+ - \vell^-) + \alpha_1 \| \vell^+ \|_{\mP}^2 + \alpha_2 \| \vell^- \|_{\mP}^2.
\end{align*}

\noindent For any sequence in the feasible set with $\|(\vell^+, \vell^-)\| \to \infty$, the quadratic terms dominate resulting in:
\begin{align*}
    f(\vell^+, \vell^-) \geq 
        \lambda_{\min}(\mP) \left( \alpha_1 \| \vell^+ \|^2 \hspace{-0.25em} + \hspace{-0.1em} \alpha_2 \| \vell^- \|^2 \right) 
        \hspace{-0.15em} - \hspace{-0.15em} \|\vk\| \| \vell^+ \hspace{-0.3em} - \hspace{-0.15em} \vell^- \|,
\end{align*}

\noindent which implies \( f(\vell^+, \vell^-) \to \infty \) as \( \|(\vell^+, \vell^-)\| \to \infty \). The indicator function \( \imath_1(\vell^+) \) is zero if \( \mathbf{1}^\top \vell^+ = -n \), and infinite otherwise. Hence, among a feasible sequence satisfying this affine equality, the value is zero and does not affect coercivity. The same applies to \( \imath_1(\vell^-) \). The indicator \( \imath_2(\vz^+, \vz^-) \) enforces that \( (\vz^+, \vz^-) \in \mathcal{H}_2 := \{ \vz^+, \vz^- \leq \mathbf{0}, \ (\vz^+)^\top \vz^- = 0 \} \). Among feasible sequences satisfying this constraint, \( \imath_2 \) remains zero. Therefore, among any feasible sequences with \( \|(\vz^+, \vz^-)\| \to \infty \), the indicator functions remain finite and the function \( f \) grows to infinity. It then follows that the objective function of our problem goes to infinity for any sequence in feasible set with $\|(\vell^+, \vell^-, \vz^+, \vz^-)\| \to \infty$, proving coercivity of the objective function over the feasible set.  

Equality constraint in the ADMM problem in \cite{wang2019global} is written as $\mA \vx + \mB \vy = \mathbf{0}$ and assumption A$2$ requires ${\rm Im}(\mA) \subseteq {\rm Im}(\mB)$. In our optimization problem in \eqref{eq:vectorized-signed-gl-admm-form}, $\mA = \mI$ and $\mB = - \mI$. Thus, assumption A$2$ is trivially satisfied. If we rewrite the optimization problem in (\ref{eq:vectorized-signed-gl-admm-form}) as:
\begin{align}
    \label{eq:convergence_algorithm}
    & \tilde{g}(\vell)+\iota_{2}(\mathbf{z}),\quad \text{such that}\ \mA \mathbf{z} + \mB \vell = \mathbf{0},\notag\\
  & \text{where}\ \vell=\binom{\vell^{+}}{\vell^{-}}\ \text {and }\ \mathbf{z}=\binom{\vz^{+}}{\vz^{-}},
\end{align}

\noindent where $\tilde{g}(\vell)= f(\vell^+, \vell^-) + \imath_1(\vell^+) + \imath_1(\vell^-)$.  Let us now define two mappings $G:\operatorname{Im}(\mB) \rightarrow \mathbb{R}^{n(n-1)}$ and $H:\operatorname{Im}(\mA) \rightarrow \mathbb{R}^{n(n-1)}$ as:
$$
G(\mathbf{u})= \argmin _{\mathbf{z} \in \mathbb{R}^{n(n-1)}}\left\{\tilde{g}(\vell)+\imath_{2}(\mathbf{z}): \mA\mathbf{ z}=\mathbf{u} \right\},
$$
$$
H(\mathbf{v})= \argmin _{\vell \in \mathbb{R}^{n(n-1)}}\left\{\tilde{g}(\vell)+\imath_{2}(\mathbf{z}): \mB\vell=\mathbf{v} \right\}.
$$
In order to show that assumption A$3$ is satisfied, we should have both minimizers $G(\mathbf{u})$ and $H(\mathbf{v})$ to be unique for each fixed $\vell$ and $\mathbf{z}$, respectively. Since
 $\mathbf{A}$ and $\mathbf{B}$ are both full column rank matrices for our case, the minimizers are unique and thus the mappings $G$ and $H$ are valid. Assumption A$3$ also requires both $G(\mathbf{u})$ and $H(\mathbf{v})$ to be Lipschitz continuous functions. We prove the Lipschitz continuity of both functions as follows:
\begin{align*}
  \|G(\mathbf{u}_1)-G(\mathbf{u}_2)\|= \|\vz_1 -\vz_2\| & \leq C_{\mathbf{u}} \|\mA\vz_1 -\mA\vz_2\|\\
  & = C_{\mathbf{u}}\|\mathbf{u}_1-\mathbf{u}_2\|, 
\end{align*}
\begin{align*}
     \|H(\mathbf{v}_1)-H(\mathbf{v}_2)\|= \|\vell_1 -\vell_2\| & \leq C_{\mathbf{v}} \|\mB\vell_1 -\mB\vell_2\|\\
  & = C_{\mathbf{v}}\|\mathbf{v}_1-\mathbf{v}_2\|. 
\end{align*}
Here both $C_{\mathbf{u}}$ and $C_{\mathbf{v}}$ are constants greater than 1 and thus we prove that both functions $G$ and $H$ are Lipschitz continuous, which implies assumption A$3$ is satisfied. 

In order to show that assumption A$4$ in \cite{wang2019global} is satisfied, we have to prove $\imath_2(\vz^+, \vz^-)$ is a lower semi-continuous function, which is guaranteed due to the closedness of the set $\mathcal{H}_{2}$. Hence, assumption A$4$ is satisfied.

Finally, to satisfy A$5$, we need to show $h(\vell^+, \vell^-) = f(\vell^+, \vell^-) + \imath_1(\vell^+) + \imath_1(\vell^-)$ is a Lipschitz differentiable function. We observe that the optimization problem will be invalid if either $\vell^+$ or $ \vell^-$ does not belong to the set $\mathcal{H}_{1}$. In that situation one of the indicator functions will take the value infinity and we are not interested in that problem. Therefore, we only consider the case where $\vell^+, \vell^- \in \mathcal{H}_1$ and thus $h(\vell^+, \vell^-) = f(\vell^+, \vell^-)$ over the set of constraints. We now show that the function $f(\vell^+, \vell^-)$ is Lipschitz differentiable.
\begin{align}
    f(\vell^+, \vell^-) = \mathbf{k}^\top(\vell^+ - \vell^-) + \alpha_1 \| \vell^+ \|_{\mathbf{P}}^2 + \alpha_2 \| \vell^- \|_{\mathbf{P}}^2
\end{align}

\noindent The first term \( \mathbf{k}^\top (\vell^+ - \vell^-) \) is linear in both \( \vell^+ \) and \( \vell^- \), and therefore differentiable everywhere with constant gradient. Hence it is Lipschitz continuous with Lipschitz constant zero. The second and third terms are quadratic functions and thus they are twice continuously differentiable and their gradients are given by
\[
\nabla_{\vell^+} \left( \alpha_1 \| \vell^+ \|_{\mathbf{P}}^2 \right) = 2\alpha_1 \mathbf{P} \vell^+, \quad
\nabla_{\vell^-} \left( \alpha_2 \| \vell^- \|_{\mathbf{P}}^2 \right) = 2\alpha_2 \mathbf{P} \vell^-.
\]
The Hessian matrices of these quadratic terms are constant and given by
$
\nabla^2_{\vell^+} f = 2\alpha_1 \mathbf{P}, 
\nabla^2_{\vell^-} f = 2\alpha_2 \mathbf{P}.
$
Since \( \mathbf{P} \) is a symmetric positive definite matrix, the eigenvalues of these Hessians are bounded, and hence the gradient maps are Lipschitz continuous. Specifically, for any two points \( \vell^+_1, \vell^+_2 \in \mathbb{R}^{n(n-1)/2} \),
\begin{align*}
    \left\| \nabla_{\vell^+} f(\vell^+_1, \cdot) - \nabla_{\vell^+} f(\vell^+_2, \cdot) \right\| & = \left\| 2\alpha_1 \mathbf{P} (\vell^+_1 - \vell^+_2) \right\|\notag \\
    & \leq 2\alpha_1 \| \mathbf{P} \|_{2} \cdot \| \vell^+_1 - \vell^+_2 \|,
\end{align*}
and similarly for \( \vell^- \). Therefore, the gradient of \( f \) is Lipschitz continuous with Lipschitz constant $  2 \max(\alpha_1, \alpha_2) \| \mathbf{P}\|_{2} $.
\qed 
\section{Proof of Theorem \ref{thrm:consistency}}
\label{appdx:consistency-proof}
The proof of Theorem \ref{thrm:consistency} relies on sub-Gaussianity of graph signals, therefore we first provide some important definitions regarding sub-Gaussian random vectors.
\begin{definition}[Sub-Gaussian Random Vectors]
\label{def:sub-gaussian}
    A random vector $\vx \in \setR^{n}$ is a sub-Gaussian vector if each of its one-dimensional linear projections exhibits sub-Gaussian behavior. More precisely, $\vx$ is considered sub-Gaussian if there exists a constant $K > 0$ such that, for any unit vector \(\mathbf{u} \in \mathbb{R}^n\):
    \begin{align}
        \setE \left[ \exp(t \vu^\top (\vx - \setE[\vx])) \right] \leq \exp(K^2 t^2/2)\ \forall t \in \setR.
    \end{align}

    \noindent Moreover, sub-Gaussian norm of $\vx$ is defined as: 
    \begin{align}
        \norm{\vx}_{\psi_2} = \sup_{\vu \in \setS^{n-1}} \norm{\vu^\top \vx}_{\psi_2},
    \end{align}

    \noindent where \(\mathbb{S}^{n-1}\) is the unit sphere in \(\mathbb{R}^n\) and $\norm{\vu^\top \vx}_{\psi_2}$ is the sub-Gaussian norm of random variable $\vu^\top \vx$ \cite{vershynin2010introduction}.
\end{definition}

If a set of random vectors are sub-Gaussian, the difference between their sample and true covariance matrix is upper bounded, as stated by the following lemma, whose proof is given in \cite{vershynin2010introduction}. 
\begin{lemma}
\label{subG-lem}
    Consider a matrix $\mX \in \setR^{n \times m}$, whose columns are assumed to be independent sub-Gaussian random vectors with second moment matrix $\mSigma$. Then for every $t \geq 0$, the following inequality holds with probability at least $1 - 2\exp(-ct^2)$:
    \begin{align*}
        \norm{\widehat{\mSigma} - \mSigma}_2 \leq \max(\delta, \delta^2) \quad \text{where} \quad \delta = C\sqrt{\frac{n}{m}} + \frac{t}{\sqrt{m}},
    \end{align*}
    where $C = C_K$, $c = c_K > 0$ depend only on the sub-Gaussian norm $K = \max_i \| \mX_{\cdot i} \|_{\psi_2}$ of the columns. 
\end{lemma}

Let $\phi(\mL^+, \mL^-)$ be the scaled version of the objective function in \eqref{eq:signed-gl} defined as follows:
\begin{align*}
    \phi(\mL^+, \mL^-) = \trace(\widehat{\mSigma}[\mL^+ - \mL^-]) + \alpha_{1m} \norm{\mL^+}_F^2 + \alpha_{2m} \norm{\mL^-}_F^2,
\end{align*}

\noindent where $\alpha_{1m} = \alpha_1/m$ and $\alpha_{2m} = \alpha_2/m$, and the first term is $\trace(\mX^\top [\mL^+ - \mL^-] \mX)/m = \trace([\mX\mX^\top/m][\mL^+ - \mL^-]) = \trace(\widehat{\mSigma}[\mL^+ - \mL^-])$. Since $\widehat{\mL}^+$ and $\widehat{\mL}^-$ are the local minimizers around a neighborhood of $\mL_0^+$ and $\mL_0^-$, we have:
\begin{align}
    & \phi(\widehat{\mL}^+, \widehat{\mL}^-) \leq \phi(\mL_0^+, \mL_0^-) \implies \notag \\[6pt]
    & \trace(\widehat{\mSigma}[\widehat{\mL}^+ - \widehat{\mL}^-]) 
        + \alpha_{1m} \norm{\widehat{\mL}^+}_F^2 
        + \alpha_{2m} \norm{\widehat{\mL}^-}_F^2 \notag \\
    & \quad \leq 
    \trace(\widehat{\mSigma}[\mL_0^+ - \mL_0^-]) 
        + \alpha_{1m} \norm{\mL_0^+}_F^2 + \alpha_{2m} \norm{\mL_0^-}_F^2 \implies \notag \\[6pt]
    & \trace(\widehat{\mSigma}[\widehat{\mL}^+ - \mL_0^+]) 
        + \alpha_{1m} (\norm{\widehat{\mL}^+}_F^2 - \norm{\mL_0^+}_F^2) \notag \\
        & \quad - \trace(\widehat{\mSigma}[\widehat{\mL}^- - \mL_0^-]) 
        + \alpha_{1m} (\norm{\widehat{\mL}^-}_F^2 - \norm{\mL_0^-}_F^2) \leq 0 \implies \notag \\[6pt]
    & \trace(\mSigma_0 [\widehat{\mL}^+ - \mL_0^+]) 
        + \alpha_{1m} (\norm{\widehat{\mL}^+}_F^2 - \norm{\mL_0^+}_F^2) \notag \\
        & \quad - \trace(\mSigma_0 [\widehat{\mL}^- - \mL_0^-]) 
        + \alpha_{1m} (\norm{\widehat{\mL}^-}_F^2 - \norm{\mL_0^-}_F^2) \leq \notag \\
    & \quad \trace([\widehat{\mSigma} - \mSigma_0] [\mL_0^+ - \widehat{\mL}^+]) 
        + \trace([\widehat{\mSigma} - \mSigma_0] [\widehat{\mL}^- - \mL_0^-]), 
        \label{eq:main-ineq}
\end{align}

\noindent where $\mSigma_0$ is the true covariance matrix. Define function $g: \setR^{n \times n} \rightarrow \setR$ such that for any $\mathbf{L} \in \mathbb{L} $ and $\alpha > 0$,\ $g(\mathbf{L}) = \alpha \left\|\mathbf{L}\right\|_{F}^{2}$. Since $g$ is a strongly convex function with parameter $2\alpha$, we have:
\begin{align}
    g(\widehat{\mL}^+) - g(\mL_0^+) & \geq 
        \trace (\nabla g(\mathbf{L}_0^{+}) [\widehat{\mL}^+ - \mL_0^+]) 
        + \alpha \norm{\widehat{\mL}^+ - \mL_0^+}_F^2, \notag \\
    \geq & - 2 \alpha \norm{\mL_0^+}_F \norm{\widehat{\mL}^+ - \mL_0^+}_F + \alpha \norm{\widehat{\mL}^+ - \mL_0^+}_F^2, \notag
\end{align}

\noindent where $\mDelta^+ = \widehat{\mL}^+ - \mL_0^+$, In the second line, we use the inequality $\trace(\mA\mB) \geq - \norm{\mA} \norm{\mB}$ for any size-compatible matrices $\mA$ and $\mB$; and $\nabla g(\mathbf{L}_0^{+}) = 2 \alpha \mL_0^+$. Applying this to Frobenius norms on the left hand side of \eqref{eq:main-ineq}, we obtain:
\begin{align}
    \alpha_{1m} (\norm{\widehat{\mL}^+}_F^2 - \norm{\mL_0^+}_F^2) \geq &
         - 2 \alpha_{1m} \norm{\mL_0^+}_F \norm{\widehat{\mL}^+ - \mL_0^+}_F \notag \\
         & + \alpha_{1m} \norm{\widehat{\mL}^+ - \mL_0^+}_F^2, \label{eq:lhs-pos-lb} \\
    \alpha_{2m} (\norm{\widehat{\mL}^-}_F^2 - \norm{\mL_0^-}_F^2) \geq &
         - 2 \alpha_{2m} \norm{\mL_0^-}_F \norm{\widehat{\mL}^- - \mL_0^-}_F \notag \\
         & + \alpha_{2m} \norm{\widehat{\mL}^- - \mL_0^-}_F^2. \label{eq:lhs-neg-lb}
\end{align}

\noindent For the first term on the right hand side of \eqref{eq:main-ineq}, we apply Cauchy-Schwarz inequality and Lemma \ref{subG-lem} with $t = \sqrt{n}$ such that we have:
\begin{align}
    \trace([\widehat{\mSigma} - \mSigma_0] [\mL_0^+ - \widehat{\mL}^+]) 
    & \leq 
        \norm{\widehat{\mSigma} - \mSigma_0}_F 
        \norm{\mL_0^+ - \widehat{\mL}^+}_F \notag \\
    & \leq \sqrt{n} \norm{\widehat{\mSigma} - \mSigma_0}_2 
        \norm{\mL_0^+ - \widehat{\mL}^+}_F \notag \\ 
    & \leq C_1 \frac{n}{\sqrt{m}} \norm{\mL_0^+ - \widehat{\mL}^+}_F, \label{eq:rhs-pos-ub}
\end{align}

\noindent holding true with probability at least $1-2\exp(-cn)$ where $C_1 = C+1$. Similarly, we have:
\begin{align}
    \trace([\widehat{\mSigma} - \mSigma_0] [\widehat{\mL}^- - \mL_0^-]) 
    \leq 
        C_1 \frac{n}{\sqrt{m}} \norm{\mL_0^- - \widehat{\mL}^-}_F, \label{eq:rhs-neg-ub}
\end{align}

\noindent The first and third terms on the left hand side of \eqref{eq:main-ineq} can be lower bounded using Cauchy-Schwarz inequality as follows:
\begin{gather}
    \trace(\mSigma_0 [\widehat{\mL}^+ - \mL_0^+]) \geq - \norm{\widehat{\mL}^+ - \mL_0^+}_F \norm{\mSigma_0}_F, \\ 
    - \trace(\mSigma_0 [\widehat{\mL}^- - \mL_0^-]) \geq - \norm{\widehat{\mL}^- - \mL_0^-}_F \norm{\mSigma_0}_F.
\end{gather}

\noindent Substituting the above two inequalities along with those in \eqref{eq:lhs-pos-lb}, \eqref{eq:lhs-neg-lb}, \eqref{eq:rhs-pos-ub} and \eqref{eq:rhs-neg-ub} into \eqref{eq:main-ineq}, we obtain:
\begin{align*}
    & - \norm{\widehat{\mL}^+ - \mL_0^+}_F \norm{\mSigma_0}_F 
        - \norm{\widehat{\mL}^- - \mL_0^-}_F \norm{\mSigma_0}_F \\ 
       & \quad + \alpha_{1m} \norm{\widehat{\mL}^+ - \mL_0^+}_F^2 
        + \alpha_{2m} \norm{\widehat{\mL}^- - \mL_0^-}_F^2 \leq \\
    & C_1 \frac{n}{\sqrt{m}} \norm{\mL_0^+ - \widehat{\mL}^+}_F + 
        C_1 \frac{n}{\sqrt{m}} \norm{\mL_0^- - \widehat{\mL}^-}_F \\
        & \quad + 2 \alpha_{1m} \norm{\mL_0^+}_F \norm{\widehat{\mL}^+ - \mL_0^+}_F 
        + 2 \alpha_{2m} \norm{\mL_0^-}_F \norm{\widehat{\mL}^- - \mL_0^-}_F.
\end{align*}

\noindent Let $\widehat{\mL}$ be the block diagonal matrix consisting of $\widehat{\mL}^+$ and $\widehat{\mL}^-$ and let $\mL_0$ be defined similarly. Letting $\alpha_{m}= \min\left\{\alpha_{1m},\alpha_{2m}\right\}$ and $\tilde{\alpha}_{m}= \max\left\{\alpha_{1m},\alpha_{2m}\right\}$, we obtain the stated inequality in Theorem \ref{thrm:consistency} as:
\begin{align}
\label{main-res}
    \norm{\widehat{\mathbf{L}}-\mathbf{L}_0}_{F} \leq C_{1} \frac{n}{\alpha_{m}\sqrt{m}} + \frac{1}{\alpha_{m}}\left(\left\|\bold{\Sigma}_0\right\|_{F} + 2 \tilde{\alpha}_{m} M  \right).
\end{align} 
\qed

\bibliographystyle{IEEEtran}
\bibliography{Reference}

\end{document}